\documentclass[11pt]{article}

\usepackage[preprint]{acl}

\usepackage{times}
\usepackage{latexsym}
\usepackage{algorithm}
\usepackage[noend]{algpseudocode}

\usepackage{subcaption} 
\usepackage[T1]{fontenc}

\usepackage[utf8]{inputenc}

\usepackage{microtype}

\usepackage{inconsolata}

\usepackage{graphicx}

\usepackage{amsmath, amssymb}
\usepackage{booktabs}
\usepackage{array}
\usepackage{multirow}

\title{Preference Heads in Large Language Models:
A Mechanistic Framework for Interpretable Personalization}

\author{%
  Weixu Zhang$^{1,2}$, Ye Yuan$^{1,2}$, Changjiang Han$^{3}$, Yuxing Tian$^{4}$, Zipeng Sun$^{1,2}$,\\  \textbf{Linfeng Du$^{1,2}$, Jikun Kang$^{5}$, Hong Kang$^{1,2}$, Xue Liu$^{1,2,3}$, Haolun Wu$^{1,2}$\thanks{Corresponding author.}}\\
  $^1$McGill University 
  $^2$Mila - Quebec AI Institute \\
  $^3$MBZUAI 
  $^4$University of Montreal
  $^5$Salesforce\\
  \texttt{\{weixu.zhang, haolun.wu\}@mail.mcgill.ca} \\
}

\begin{document}
\maketitle

\begin{abstract}

Large Language Models (LLMs) exhibit strong implicit personalization ability, yet most existing approaches treat this behavior as a black box, relying on prompt engineering or fine tuning on user data. In this work, we adopt a mechanistic interpretability perspective and hypothesize the existence of a sparse set of \textbf{Preference Heads}, attention heads that encode user specific stylistic and topical preferences and exert a causal influence on generation. We introduce \textbf{Differential Preference Steering (DPS)}, a training free framework that (1) identifies Preference Heads through causal masking analysis and (2) leverages them for controllable and interpretable personalization at inference time. DPS computes a \textbf{Preference Contribution Score (PCS)} for each attention head, directly measuring its causal impact on user aligned outputs. During decoding, we contrast model predictions with and without Preference Heads, amplifying the difference between personalized and generic logits to selectively strengthen preference aligned continuations. Experiments on widely used personalization benchmarks across multiple LLMs demonstrate consistent gains in personalization fidelity while preserving content coherence and low computational overhead. Beyond empirical improvements, DPS provides a mechanistic explanation of where and how personalization emerges within transformer architectures. Our implementation is publicly available.\footnote{\url{https://github.com/weixuzhang/DPS}}

\end{abstract}

\section{Introduction}

Large Language Models (LLMs) exhibit strong implicit personalization ability.
They often adapt to a user-specific tone, writing style, and topical interests with minimal conditioning.
This capability is critical for applications such as personalized writing assistance, recommendation generation, and conversational agents, where users expect outputs to reflect individual preferences.
Recent surveys highlight rapid progress in personalized LLMs via prompting, retrieval, and fine-tuning-based approaches \cite{DBLP:journals/corr/abs-2502-11528}.
However, most existing methods treat LLMs as black-box systems.
They rely on prompt engineering, retrieval-based user profiles, or fine-tuning on user data \cite{DBLP:conf/acl/SalemiMBZ24,DBLP:journals/corr/abs-2407-11016,DBLP:conf/iclr/ChenZLCL25}, without understanding how user preferences are internally represented or propagated.
As a result, these approaches often improve surface-level consistency but lack interpretability, controllability, and scalability across users.

\begin{figure}[t]
  \centering
  \includegraphics[width=\columnwidth]{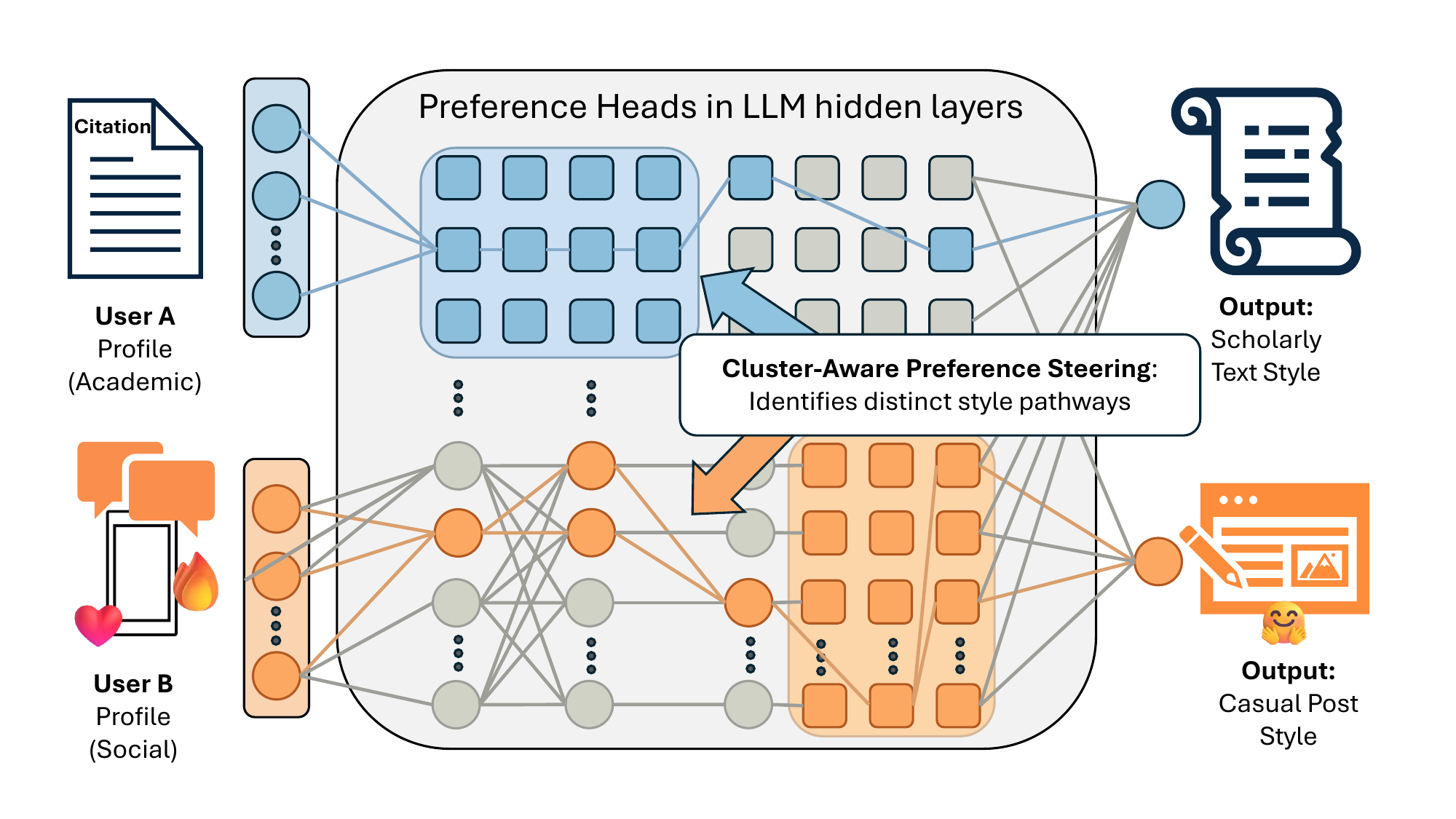}
  \caption{Overview of preference-based personalization in LLMs.
Distinct user profiles activate different subsets of \emph{Preference Heads}, forming sparse internal pathways that steer generation toward user-aligned styles.
Cluster-aware preference steering further captures shared structure across users with similar preferences.}
  \label{fig:introduction}
  \vspace{-0.5cm}
\end{figure}

Recent advances in mechanistic interpretability suggest that many transformer behaviors can be traced to small, specialized internal components.
Prior work has identified attention heads responsible for sequence copying, long-range retrieval, and abstract task representations, showing that high-level capabilities may emerge from localized circuits rather than diffuse parameter interactions \cite{DBLP:conf/iclr/WuWX0F25,DBLP:journals/corr/abs-2506-09944,DBLP:conf/icml/YinS25}.
Related studies further demonstrate that attention heads can act as efficient zero-shot re-rankers or encode structured interaction patterns \cite{DBLP:conf/iclr/ChenGS25,DBLP:conf/acl/QuYLWLDC25,reattn}.
These findings raise a natural but underexplored question: \emph{where does personalization arise within large language models?}

In this work, we argue that personalization is mediated by a sparse subset of attention heads that encode user-specific stylistic and topical signals and exert a direct influence on generation.
We refer to these components as \textbf{Preference Heads}.
As illustrated in Figure~\ref{fig:introduction}, different users activate distinct internal pathways through these heads, leading to systematically different generation behaviors (e.g., scholarly versus conversational styles).
Unlike general semantic or retrieval-oriented heads, Preference Heads selectively amplify tokens that are consistent with a user profile.
While prior work has explored preference learning from user edits or feedback \cite{DBLP:conf/icml/TuckerBCJ24,DBLP:conf/nips/PangYHCSW24}, it does not examine how preferences are represented at the level of internal model components.

To identify Preference Heads, we introduce a causal head discovery procedure based on targeted masking and contribution analysis.
We define a \textbf{Preference Contribution Score (PCS)} that measures the causal impact of each attention head on user-aligned generation.
This approach builds on recent mechanistic analyses of attention heads for retrieval and factuality \cite{DBLP:conf/iclr/WuWX0F25,DBLP:journals/corr/abs-2410-18860}, but focuses specifically on personalization.

Building on this analysis, we propose \textbf{Differential Preference Steering (DPS)}, a decoding-time framework that enables interpretable and controllable personalization without modifying model parameters.
DPS contrasts model predictions with and without Preference Heads and amplifies their difference during decoding, strengthening preference-aligned continuations while suppressing generic ones.
To account for user heterogeneity, we further extend DPS with cluster-aware preference steering, allowing users with similar profiles to share partially overlapping preference circuits.
While related to contrastive decoding and activation steering methods \cite{DBLP:journals/corr/abs-2410-18860,DBLP:conf/acl/Zhang0WLWWC25,DBLP:conf/acl/WangXMDTC025}, DPS targets causally identified personalization circuits rather than externally defined behaviors.

We evaluate DPS on widely used personalization benchmarks covering both short-form and long-form user-conditioned generation \cite{DBLP:conf/acl/SalemiMBZ24,DBLP:journals/corr/abs-2407-11016}.
Across multiple open-source LLMs, DPS consistently improves personalization fidelity and stylistic alignment while preserving content coherence and computational efficiency.
Further analyses show that Preference Heads are causally meaningful and interpretable: masking them disrupts personalized behavior, and their activation patterns correspond to structured stylistic and topical dimensions.

Our key contributions are as follows:
\begin{itemize}
    \item We present the first mechanistic analysis of personalization in LLMs by identifying \textbf{Preference Heads}, a sparse set of attention heads that causally encode user-specific preferences.
    \item We introduce \textbf{Preference Contribution Score} and \textbf{Differential Preference Steering}, a training-free framework for interpretable and controllable personalization at decoding time.
    \item Through extensive experiments and analyses, we demonstrate that targeted intervention on Preference Heads yields consistent personalization gains across model architectures, offering a principled alternative to black-box personalization methods.
\end{itemize}

\begin{figure*}[t]
  \centering
  \includegraphics[width=0.95\linewidth]{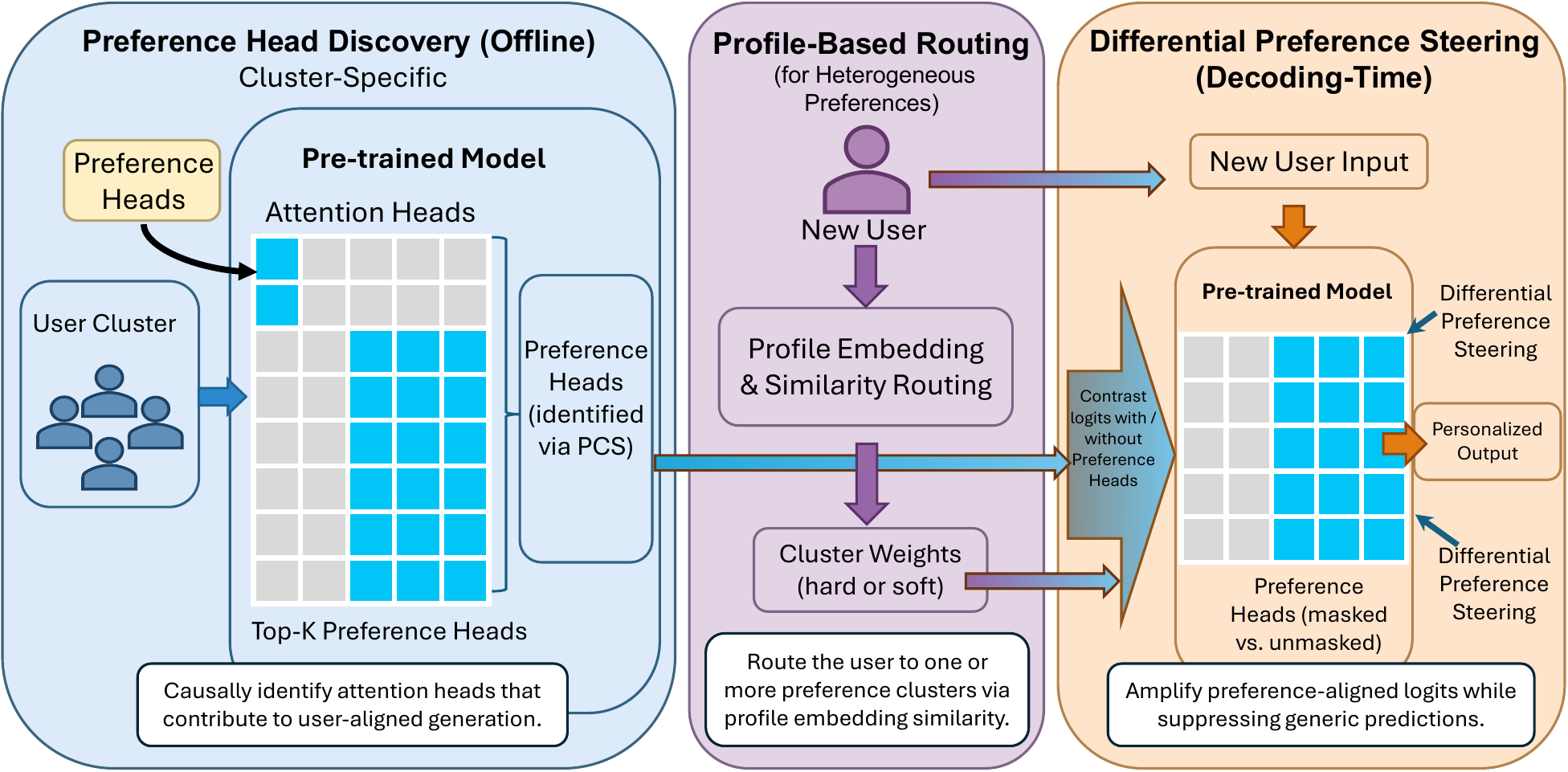}
  \caption{Overview of the proposed framework.
We first perform offline, causal discovery of Preference Heads by measuring the effect of head ablation on user-aligned generation.
For heterogeneous users, profiles can optionally be clustered and routed via embedding similarity.
At decoding time, Differential Preference Steering contrasts model logits with and without Preference Heads and amplifies preference-aligned differences to produce personalized outputs without parameter updates.}
  \label{fig:framework}
  \vspace{-0.3cm}
\end{figure*}

\section{Related Work}

\subsection{Personalization in Large Language Models}

Personalization in large language models aims to adapt generation behavior to user specific preferences such as writing style, topical interests, or decision patterns.
Prior work explores fine tuning, prompting, retrieval, and decoding time methods to achieve personalization, with benchmarks such as LaMP and LongLaMP enabling systematic evaluation \cite{DBLP:conf/acl/SalemiMBZ24,adaptivepersonalizedcir}.
Several approaches model preferences explicitly, including personalized decoding \cite{DBLP:conf/iclr/ChenZLCL25}, difference aware user modeling \cite{DBLP:conf/acl/Qiu00B00FC25}, and latent preference learning from user edits or feedback \cite{DBLP:conf/icml/TuckerBCJ24,DBLP:conf/nips/GaoTSMM24}.
While effective, these methods treat personalization as an external conditioning problem and do not examine how user preferences are internally represented within transformer components.

\subsection{Mechanistic Interpretability and Attention Heads}

Mechanistic interpretability studies show that transformer behavior can often be attributed to specialized internal components.
Induction heads support pattern matching in context \cite{DBLP:conf/naacl/CrosbieS25}, retrieval heads enable long context factual recall \cite{DBLP:conf/iclr/WuWX0F25,DBLP:journals/corr/abs-2506-09944}, and other attention heads act as efficient rerankers or encode structured interactions \cite{DBLP:conf/iclr/ChenGS25,DBLP:conf/acl/QuYLWLDC25}.
Related work also examines which attention heads matter for specific tasks and shows that a sparse subset of internal units can dominate model behavior \cite{DBLP:conf/icml/YinS25,DBLP:conf/acl/Jiao0TMPA24}.
However, existing analyses primarily focus on reasoning, factuality, or in context learning rather than personalization.

\subsection{Decoding Time Control and Contrastive Methods}

Decoding time control provides a model agnostic way to steer language model behavior without retraining \cite{opendecoder}.
Contrastive approaches guide generation by amplifying differences between model variants, including layer level contrast in DoLa \cite{DBLP:conf/iclr/ChuangXLKGH24} and head level contrast in DeCoRe \cite{DBLP:journals/corr/abs-2410-18860}.
Related methods apply activation steering or contrastive decoding to influence generation style or behavior \cite{DBLP:conf/acl/Zhang0WLWWC25,DBLP:conf/acl/WangXMDTC025}.
Although some work applies decoding time control to personalization \cite{DBLP:conf/iclr/ChenZLCL25}, they do not identify the internal components responsible for personalization, whereas our approach grounds decoding time intervention in a mechanistic analysis of attention heads.

\section{Preference Head Discovery}
\label{sec:preference-head-discovery}

In this section, we formalize the notion of \emph{Preference Heads} and introduce a causal procedure for identifying them.
Our goal is to determine which attention heads inside a pretrained transformer are directly responsible for injecting user specific preferences into the generation process.

\subsection{Problem Setup and Intuition}

We consider a pretrained transformer based language model parameterized by $\theta$.
Given a task input $x$ and a user profile $u$, the model generates an output sequence $y = (y_1, \ldots, y_T)$ autoregressively.
At decoding step $t$, the model produces a distribution over the vocabulary
\begin{equation}
p_\theta(y_t \mid x, u, y_{<t}) = \mathrm{Softmax}(\ell_t),
\end{equation}
where $\ell_t \in \mathbb{R}^{|\mathcal{V}|}$ denotes the output logits.

The logits $\ell_t$ reflect a mixture of generic linguistic knowledge and user specific preference signals.
While existing personalization methods influence this mixture externally through prompting or fine tuning, we seek to identify where preference information is internally encoded.
Motivated by prior mechanistic analyses, we hypothesize that personalization is mediated by a sparse subset of attention heads whose activations causally affect user aligned generation.
We refer to these components as \emph{Preference Heads}.

\begin{algorithm*}[ht]
\caption{Differential Preference Steering with Heterogeneous Preferences}
\label{alg:dps}
\hspace*{\algorithmicindent} \textbf{Input:}
User-conditioned dataset $\mathcal{D} = \{(x_i, u_i, y_i^\ast)\}_{i=1}^N$,
context $x$, user profile $u$,
pretrained model $M_\theta$,
personalization strength $\gamma$ \\
\hspace*{\algorithmicindent} \textbf{Output:}
Personalized output sequence $y = (y_1, \ldots, y_T)$
\begin{algorithmic}[1]

\State \texttt{\textcolor{gray}{/* Phase 1: Preference Head Discovery */}}
\For{each attention head $h_{l,k}$ in $M_\theta$}
    \State Construct ablated model $M_{\theta \setminus h_{l,k}}$
    \State Compute
    $
    \mathrm{PCS}(h_{l,k})
    \leftarrow
    \mathbb{E}_{(x,u,y^\ast)\sim\mathcal{D}}
    \big[
    \mathcal{L}(M_{\theta \setminus h_{l,k}}, x, u, y^\ast)
    -
    \mathcal{L}(M_\theta, x, u, y^\ast)
    \big]
    $
\EndFor
\State Select top $K$ heads with highest PCS values
\State $\mathcal{H}_{\mathrm{pref}} \leftarrow \mathrm{TopK}(\mathrm{PCS})$

\vspace{0.6em}
\State \texttt{\textcolor{gray}{/* Phase 2: Clustering and Routing */}}
\State Compute profile embeddings for all users
\State Cluster users into groups $\{c\}$ using embedding similarity
\State Assign user $u$ to cluster(s) with weight(s) $\{w_c\}$ \Comment{hard or soft routing}
\For{each cluster $c$}
    \State Obtain cluster-specific Preference Heads $\mathcal{H}_{\mathrm{pref}}^{(c)}$
\EndFor

\vspace{0.6em}
\State \texttt{\textcolor{gray}{/* Phase 3: Differential Preference Steering */}}
\State Initialize decoding context $z \leftarrow (x, u)$
\For{$t = 1$ to $T$}
    \State Compute personalized logits
    $
    \ell_t^{\mathrm{pref}} \leftarrow f_{M_\theta}(z)
    $
    \State Compute depersonalized logits with weighted head suppression
    $
    \ell_t^{\mathrm{gen}} \leftarrow
    f_{M_{\theta \setminus \sum_c w_c \mathcal{H}_{\mathrm{pref}}^{(c)}}}(z)
    $
    \State Combine logits
    $
    \tilde{\ell}_t \leftarrow (1 + \gamma)\,\ell_t^{\mathrm{pref}} - \gamma\,\ell_t^{\mathrm{gen}}
    $
    \State Sample or select next token
    $
    y_t \sim \mathrm{Softmax}(\tilde{\ell}_t)
    $
    \State Update context $z \leftarrow (z, y_t)$
\EndFor
\State \Return $y$
\end{algorithmic}
\end{algorithm*}

\subsection{Causal Identification via Head Ablation}

Let $h_{l,k}$ denote the $k$th attention head in layer $l$.
To assess whether a head contributes to personalization, we perform targeted ablation by masking its output during inference.
Let $M_\theta$ denote the original model and $M_{\theta \setminus h_{l,k}}$ denote the model with head $h_{l,k}$ masked. Given a dataset of user conditioned examples:
\begin{equation}
\mathcal{D} = \{(x_i, u_i, y_i^\ast)\}_{i=1}^N,
\end{equation}
we measure how ablating $h_{l,k}$ affects the likelihood of the reference output $y_i^\ast$.

For a single example, we define the negative log likelihood
\begin{equation}
\mathcal{L}(M, x, u, y^\ast)
= - \frac{1}{T} \sum_{t=1}^{T}
\log p_M(y_t^\ast \mid x, u, y_{<t}^\ast).
\end{equation}
If masking a head increases this loss, it indicates that the head contributes causally to user aligned generation behavior.

\subsection{Preference Contribution Score}

Based on this intuition, we define the \emph{Preference Contribution Score} (PCS) for each attention head.
For head $h_{l,k}$, PCS is computed as the expected loss difference between the original model and the ablated model:
\begin{align}
\mathrm{PCS}(h_{l,k})
&=
\mathbb{E}_{(x,u,y^\ast)\sim \mathcal{D}}
\big[
\mathcal{L}(M_{\theta \setminus h_{l,k}}, x, u, y^\ast)
\nonumber \\
&\quad -
\mathcal{L}(M_\theta, x, u, y^\ast)
\big].
\end{align}

A positive PCS indicates that removing the head degrades user aligned generation, suggesting a causal contribution to personalization.
Because PCS is computed through direct intervention, it measures causal effect rather than correlation.

\subsection{Selecting Preference Heads}

\begin{table*}[!ht]
\small
\centering
\resizebox{\textwidth}{!}{
\begin{tabular}{llccccccccc}
\toprule
\multirow{2}{*}{\textbf{Model}} &
\multirow{2}{*}{\textbf{Method}} &
\multicolumn{3}{c}{\textbf{News Headline Generation}} &
\multicolumn{3}{c}{\textbf{Scholarly Title Generation}} &
\multicolumn{3}{c}{\textbf{Tweet Paraphrasing}} \\
\cmidrule(lr){3-5} \cmidrule(lr){6-8} \cmidrule(lr){9-11}
& 
& \textbf{R-1}~$\uparrow$ & \textbf{R-L}~$\uparrow$ & \textbf{METEOR}~$\uparrow$
& \textbf{R-1}~$\uparrow$ & \textbf{R-L}~$\uparrow$ & \textbf{METEOR}~$\uparrow$
& \textbf{R-1}~$\uparrow$ & \textbf{R-L}~$\uparrow$ & \textbf{METEOR}~$\uparrow$ \\
\midrule

\multirow{4}{*}{LLaMA-3-8B}
& CAD
& 0.1681 & 0.1498 & 0.1568
& 0.3530 & 0.3068 & 0.3925
& 0.3368 & 0.2893 & 0.2813 \\
& DeCoRe
& 0.1768 & 0.1572 & 0.1626
& \textbf{0.4010} & \textbf{0.3527} & 0.4004
& 0.3231 & 0.2764 & 0.2729 \\
& DoLa
& 0.1694 & 0.1508 & 0.1592
& 0.3636 & 0.3117 & \textbf{0.4079}
& 0.3365 & 0.2877 & 0.2795 \\
\cmidrule(r){2-11}
& DPS (ours)
& \textbf{0.1787} & \textbf{0.1596} & \textbf{0.1650}
& 0.3243 & 0.2787 & 0.3826
& \textbf{0.3389} & \textbf{0.2898} & \textbf{0.2884} \\
\midrule

\multirow{4}{*}{Qwen2-7B}
& CAD
& 0.1580 & 0.1392 & 0.1255
& 0.4197 & \textbf{0.3780} & 0.4381
& \textbf{0.3590} & \textbf{0.3106} & \textbf{0.3384} \\
& DeCoRe
& 0.1581 & 0.1305 & 0.1232
& \textbf{0.4311} & 0.3729 & 0.4565
& 0.3470 & 0.3065 & 0.3173 \\
& DoLa
& 0.1642 & 0.1473 & 0.1272
& 0.4277 & 0.3746 & \textbf{0.4596}
& 0.3524 & 0.3046 & 0.3246 \\
\cmidrule(r){2-11}
& DPS (ours)
& \textbf{0.1627} & \textbf{0.1450} & \textbf{0.1318}
& 0.4071 & 0.3421 & 0.4230
& 0.3533 & 0.2981 & 0.3269 \\
\midrule

\multirow{4}{*}{Mistral-7B}
& CAD
& 0.1361 & 0.1132 & 0.0980
& \textbf{0.4375} & 0.3712 & 0.4561
& 0.3342 & 0.2916 & \textbf{0.3097} \\
& DeCoRe
& 0.1299 & 0.1085 & 0.0908
& 0.4135 & 0.3648 & 0.4419
& 0.3407 & 0.2927 & 0.2978 \\
& DoLa
& 0.1362 & 0.1136 & 0.0962
& 0.4364 & \textbf{0.3733} & \textbf{0.4605}
& 0.3291 & 0.2852 & 0.3026 \\
\cmidrule(r){2-11}
& DPS (ours)
& \textbf{0.1536} & \textbf{0.1366} & \textbf{0.1399}
& 0.3983 & 0.3350 & 0.4162
& \textbf{0.3441} & \textbf{0.2998} & 0.2990 \\
\bottomrule
\end{tabular}}
\caption{Results on LaMP generation tasks. $\uparrow$ indicates higher is better. Best results per model are in bold.}
\label{tab:lamp_gen}
\end{table*}

We compute PCS for all attention heads and select a small subset with the highest scores as Preference Heads.
Formally, let $\mathcal{H}$ denote the set of all attention heads and define
\begin{equation}
\mathcal{H}_{\mathrm{pref}} =
\mathrm{TopK}_{h \in \mathcal{H}} \ \mathrm{PCS}(h),
\end{equation}
where $K$ is chosen such that $\mathcal{H}_{\mathrm{pref}}$ remains sparse relative to the total number of heads.

\section{Differential Preference Steering}
\label{sec:dps}

In this section, we introduce \emph{Differential Preference Steering} (DPS), a decoding time method that leverages discovered Preference Heads to enable controllable and interpretable personalization.
DPS modifies the generation process by isolating and amplifying the internal preference signal contributed by Preference Heads, without changing model parameters or requiring additional training.

\subsection{Decomposing Generic and Preference Conditioned Behavior}

At each decoding step $t$, the output logits $\ell_t$ reflect the combined influence of multiple internal components.
Based on the analysis in Section~\ref{sec:preference-head-discovery}, we view these logits as consisting of a generic component that captures task relevant and linguistic information, and a preference conditioned component that reflects user specific stylistic and topical tendencies.

Let $M_\theta$ denote the original model and let $M_{\theta \setminus \mathcal{H}_{\mathrm{pref}}}$ denote the model with Preference Heads masked.
For a given context $(x, u, y_{<t})$, we define
\begin{align}
\ell_t^{\mathrm{pref}} &= f_{M_\theta}(x, u, y_{<t}), \\
\ell_t^{\mathrm{gen}} &= f_{M_{\theta \setminus \mathcal{H}_{\mathrm{pref}}}}(x, u, y_{<t}),
\end{align}
where $f_M(\cdot)$ denotes the logit function of $M$.

\subsection{Differential Logit Combination}

DPS constructs modified logits by amplifying the contrast between preference conditioned and generic predictions:
\begin{equation}
\tilde{\ell}_t = (1 + \gamma)\, \ell_t^{\mathrm{pref}} - \gamma \, \ell_t^{\mathrm{gen}},
\end{equation}
where $\gamma \geq 0$ controls personalization strength.

The next token distribution is given by
\begin{equation}
p(y_t \mid x, u, y_{<t}) = \mathrm{Softmax}(\tilde{\ell}_t),
\end{equation}
and decoding proceeds autoregressively using standard decoding strategies.

\section{Extending to Heterogeneous Preferences}
\label{sec:heterogeneous}

Preference Heads are not universal.
Different users may rely on different internal circuits depending on their stylistic and topical tendencies.
To account for this heterogeneity, we extend our framework with cluster aware Preference Head discovery and weighted DPS.

\subsection{Profile Embedding and Clustering}

We represent each user profile as a dense embedding computed from historical text using a pretrained sentence encoder.
Users are clustered based on embedding similarity using k means, with the number of clusters chosen to ensure sufficient samples per group.
This clustering captures coarse preference types such as formal versus conversational style or domain specific interests.

\subsection{Cluster Specific Preference Heads}

For each cluster, we repeat the PCS based head discovery procedure using only samples from that cluster.
This yields cluster specific Preference Head sets and corresponding importance weights.
Heads that are important for one cluster may be less important for others, reflecting diversity in personalization mechanisms.

\begin{table*}[!ht]
\small
\centering
\resizebox{\textwidth}{!}{
\begin{tabular}{llcccccc}
\toprule
\multirow{2}{*}{\textbf{Model}} &
\multirow{2}{*}{\textbf{Method}} &
\multicolumn{2}{c}{\textbf{Citation Identification}} &
\multicolumn{2}{c}{\textbf{Movie Tagging}} &
\multicolumn{2}{c}{\textbf{Product Rating}} \\
\cmidrule(lr){3-4} \cmidrule(lr){5-6} \cmidrule(lr){7-8}
& 
& \textbf{Accuracy}~$\uparrow$ & \textbf{F1}~$\uparrow$
& \textbf{Accuracy}~$\uparrow$ & \textbf{F1}~$\uparrow$
& \textbf{MAE}~$\downarrow$ & \textbf{RMSE}~$\downarrow$ \\
\midrule

\multirow{5}{*}{LLaMA-3-8B}
& CAD        & 0.6240 & 0.6070 & 0.4552 & 0.3839 & 0.4426 & 0.9300 \\
& DeCoRe     & 0.6232 & 0.6200 & \textbf{0.4639} & \textbf{0.4034} & 0.4442 & 0.9458 \\
& DoLa       & 0.6156 & 0.5961 & 0.2800 & 0.1643 & \textbf{0.4200} & \textbf{0.8718} \\
\cmidrule(r){2-8}
& DPS (ours) & \textbf{0.6356} & \textbf{0.6288} & 0.4610 & 0.3910 & 0.4236 & 0.9278 \\
\midrule

\multirow{4}{*}{Qwen2-7B}
& CAD        & 0.6230 & 0.6250 & 0.1850 & 0.1181 & \textbf{0.3180} & \textbf{0.6240} \\
& DeCoRe     & 0.5400 & 0.5891 & 0.2320 & 0.1217 & 0.3250 & 0.6325 \\
& DoLa       & 0.6790 & 0.6795 & 0.2412 & 0.0958 & 0.3200 & 0.6300 \\
\cmidrule(r){2-8}
& DPS (ours) & \textbf{0.6932} & \textbf{0.7078} & \textbf{0.3902} & \textbf{0.3202} & 0.3276 & 0.6719 \\
\bottomrule
\end{tabular}}
\caption{Results on LaMP classification and regression tasks. $\uparrow$ indicates higher is better and $\downarrow$ indicates lower is better. Best results per model are in bold.}
\label{tab:lamp_cls}
\end{table*}

\subsection{Weighted Differential Preference Steering}

At inference time, we route each user to clusters based on profile similarity, using either hard assignment to the nearest cluster or soft assignment that distributes weight across multiple clusters.
We aggregate cluster specific head weights into a single weighted importance map, which determines how strongly each head is suppressed in the depersonalized pass.
Weighted DPS then applies contrastive decoding using these soft head suppressions, enabling smooth personalization control across heterogeneous users.

Algorithm~\ref{alg:dps} provides a complete description of Differential Preference Steering with support for heterogeneous user preferences.

\section{Experiments}

\subsection{Experimental Setup}

\paragraph{Models}
We evaluate Differential Preference Steering (DPS) on multiple instruction tuned open source large language models:
LLaMA-3-8B-Instruct,
Mistral-7B-Instruct,
and Qwen2-7B-Instruct.
These models represent different architectural and training design choices while maintaining comparable parameter scales.

\paragraph{Datasets}
We conduct experiments on the \textbf{LaMP} benchmark~\cite{DBLP:conf/acl/SalemiMBZ24}, which evaluates personalized language modeling across both generation and prediction tasks.
LaMP includes user conditioned text generation tasks (e.g., headline generation, scholarly title generation, tweet paraphrasing) as well as classification and regression tasks (e.g., citation identification, movie tagging, product rating).
We follow the official data splits and evaluation protocols.

\paragraph{Metrics}
We report task appropriate automatic metrics.
For generation tasks, we use ROUGE-1, ROUGE-L~\cite{lin-2004-rouge}, and METEOR~\cite{DBLP:conf/acl/BanerjeeL05}.
For classification tasks, we report Accuracy and F1.
For regression tasks, we report Mean Absolute Error (MAE) and Root Mean Squared Error (RMSE).

\paragraph{Baselines}
We compare DPS against strong decoding time intervention baselines:
\textbf{Context Aware Decoding (CAD)}~\cite{DBLP:conf/naacl/ShiHLTZY24},
\textbf{Decoding by Contrasting Layers (DoLa)}~\cite{DBLP:conf/iclr/ChuangXLKGH24},
and \textbf{Decoding by Contrasting Retrieval Heads (DeCoRe)}~\cite{DBLP:journals/corr/abs-2410-18860}.
All baselines are implemented using their publicly released code or faithful reimplementations, with hyperparameters tuned following the original papers.

\subsection{Results}

\begin{figure*}[t]
\centering
\includegraphics[width=\textwidth]{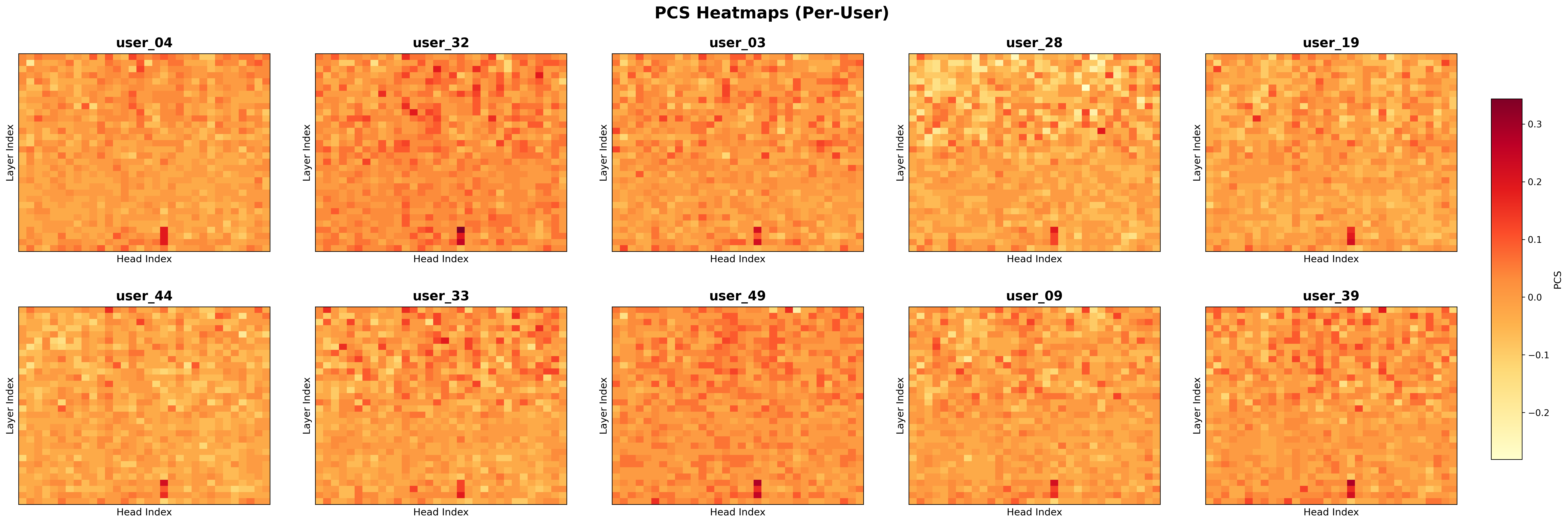}
\caption{Per-user PCS heatmaps across layers and attention heads. Users are selected randomly. Preference Heads are sparse and causally significant within users.}
\label{fig:pcs_heatmap}
\vspace{-0.35cm}
\end{figure*}

\begin{figure*}[t]
\centering
\includegraphics[width=0.7\textwidth]{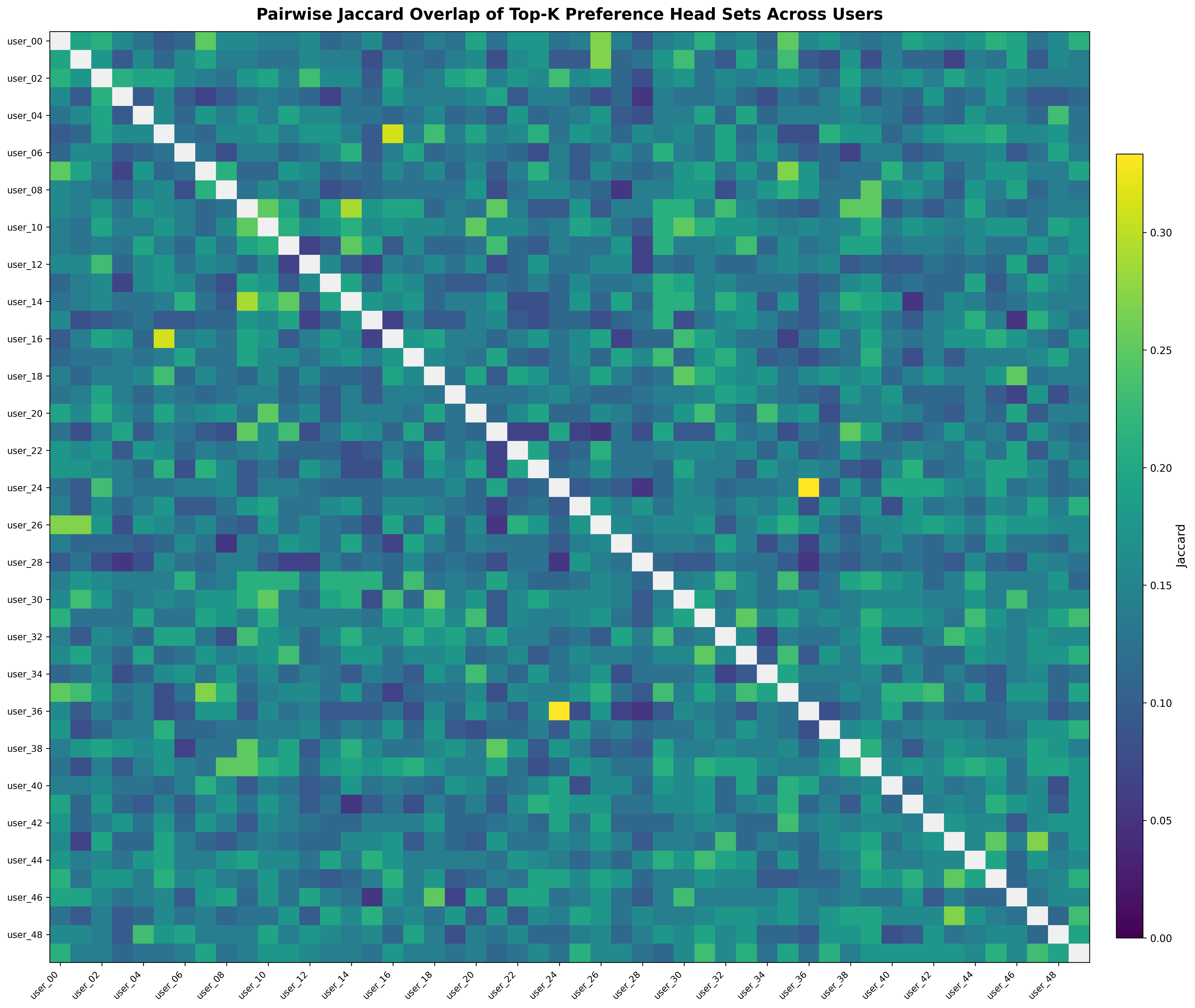}
\caption{Pairwise Jaccard overlap of top-$K$ Preference Head sets across users. Preference Heads exhibit limited overlap across users, motivating cluster-aware head discovery.}
\label{fig:user_jaccard}
\vspace{-0.35cm}
\end{figure*}

Tables~\ref{tab:lamp_gen} and~\ref{tab:lamp_cls} report results on LaMP generation, classification, and regression tasks.
Across models and tasks, DPS consistently achieves strong or best performance compared to existing decoding-time baselines.

On generation tasks, DPS improves ROUGE and METEOR scores on News Headline Generation and Tweet Paraphrasing across all evaluated models.
While some baselines achieve strong performance on individual tasks, DPS demonstrates more consistent gains across task types, indicating improved personalization fidelity without sacrificing content quality.

On classification and regression tasks, DPS achieves the best or near-best performance in most settings.
In particular, DPS substantially improves performance on Movie Tagging for Qwen2-7B and achieves the lowest error on Product Rating tasks compared to contrastive decoding baselines.
These results suggest that amplifying preference-specific internal signals benefits both generative and predictive personalization tasks.

Overall, DPS delivers consistent gains across model families and task formats, suggesting that preference head-based decoding is a robust and broadly applicable personalization mechanism rather than a model- or task-specific effect.

\begin{figure*}[!ht]
\centering
\includegraphics[width=\textwidth]{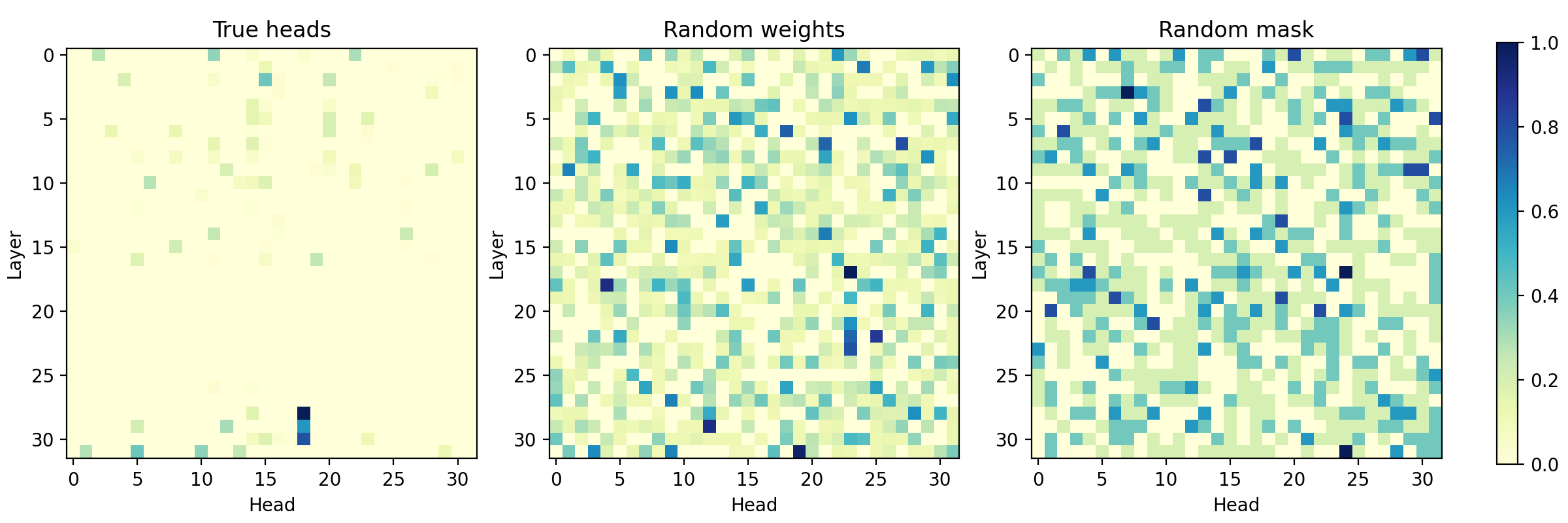}
\vspace{-0.1cm}
\caption{Ablation analysis of attention head selection.
Discovered Preference Heads exhibit sparse and structured importance (left),
while random weights and random masks (middle, right) produce diffuse patterns and fail to recover comparable personalization behavior.}
\label{fig:ablation_heads}
\vspace{-0.4cm}
\end{figure*}

\subsection{Head Discovery Analysis}
\label{sec:head_analysis}

We analyze the structure and consistency of Preference Heads identified by the Preference Contribution Score (PCS).

\paragraph{Sparse and user-specific Preference Heads.}
Per-user PCS heatmaps (Figure~\ref{fig:pcs_heatmap}) reveal that high-PCS heads are sparse and unevenly distributed across layers.
While a small number of heads consistently exhibit strong causal influence for each user, the locations of these heads vary substantially across users.
This indicates that personalization is driven by localized internal components rather than uniformly distributed attention behavior.

\paragraph{Limited cross-user overlap.}
To quantify consistency across users, we compute pairwise Jaccard overlap between the top-$K$ Preference Head sets identified independently for each user.
As shown in Figure~\ref{fig:user_jaccard}, overlap between different users remains low, with most values clustered near zero.
This suggests that Preference Heads are not shared globally, and that directly aggregating per-user head sets would dilute preference-specific signals.

\paragraph{Implications.}
Together, these results show that Preference Heads are causally meaningful but heterogeneous across users.
This motivates the cluster-aware extension introduced in Section~\ref{sec:heterogeneous}, which stabilizes head discovery by sharing preference signals among users with similar profiles.

\begin{table}[!t]
\small
\centering
\resizebox{\columnwidth}{!}{
\begin{tabular}{lccc}
\toprule
\textbf{Prompt Length} & \textbf{Baseline} & \textbf{DPS} & \textbf{Overhead} \\
\midrule
512  & 6.57 & 6.96 & 1.06$\times$ \\
1024 & 13.04 & 13.43 & 1.03$\times$ \\
2048 & 26.80 & 27.21 & 1.02$\times$ \\
\bottomrule
\end{tabular}}
\caption{Estimated inference FLOPs (TFlop) for LLaMA-3-8B with $T=32$ generated tokens.
DPS shares prompt prefill cost with standard decoding and adds a second decoding pass.}
\label{tab:efficiency}
\vspace{-0.3cm}
\end{table}

\subsection{Ablation Studies}
\label{sec:ablation}

We conduct ablation studies to validate that the gains of DPS arise from causally identified Preference Heads rather than arbitrary head manipulation.

\paragraph{True heads vs random controls.}
We compare DPS using discovered Preference Heads against two control variants:
(i) random head selection with the same cardinality, and
(ii) random masking patterns with matched sparsity.
Figure~\ref{fig:ablation_heads} visualizes the resulting head importance maps.
In contrast to the sparse and localized structure of true Preference Heads, random controls produce diffuse and unstructured patterns.

Quantitatively, replacing Preference Heads with random heads consistently degrades performance across tasks, while random masking yields unstable and task-dependent behavior.
These results confirm that DPS relies on semantically meaningful internal components rather than generic sparsification effects.

\paragraph{Effect of head selection strategy.}
We further evaluate performance when progressively increasing the number of selected heads.
As shown in Figure~\ref{fig:heads_sweep} in appendix, performance improves when including more high-PCS heads, but saturates beyond a moderate head count.
This indicates that personalization signals are concentrated in a limited subset of heads, and including too many heads introduces noise rather than additional benefit.

\subsection{Hyperparameter Sensitivity}
\label{sec:hyperparam}

We analyze the sensitivity of DPS to key design choices, focusing on the number of Preference Heads $K$ and the routing strategy used in cluster-aware personalization. Additional analysis is provided in Appendix~\ref{sec:appendix_hyperparam}.

\paragraph{Number of Preference Heads.}
Figure~\ref{fig:heads_sweep} shows performance as a function of the number of selected heads.
DPS remains stable across a wide range of $K$, with peak performance achieved using a moderate number of heads.
This robustness suggests that precise tuning of $K$ is not critical, as long as the selection focuses on high-PCS heads.

\paragraph{Head set stability across scales.}
To assess how head sets evolve as $K$ increases, we compute pairwise Jaccard overlap between head sets selected at different values of $K$.
As shown in Figure~\ref{fig:headset_jaccard}, head sets become increasingly stable at larger $K$, indicating that top-ranked heads are consistently preserved while additional heads contribute diminishing returns.

\paragraph{Routing strategy.}
Finally, we compare hard routing and soft routing for cluster-aware DPS.
Figure~\ref{fig:routing_ablation} in the appendix shows that hard routing performs slightly better on classification tasks, while soft routing provides more robust gains on generation tasks.
This trade-off suggests that soft routing offers smoother personalization control, whereas hard routing emphasizes stronger specialization.

\subsection{Efficiency Analysis}

We analyze the inference-time computational cost of Differential Preference Steering (DPS).
DPS is training-free and introduces no additional parameters or prompt modifications.

\paragraph{Inference Cost}
At decoding time, DPS performs two forward passes per step: one using the original model and one with Preference Heads masked.
This cost is comparable to existing contrastive decoding methods such as CAD and DeCoRe.
Unlike CAD, DPS does not require re-encoding modified prompts and therefore incurs no additional prefill overhead.
Since prompt prefill typically dominates the inference cost for realistic context lengths, the relative overhead of our DPS decreases as prompt length increases.

\paragraph{Computation Cost}
Table~\ref{tab:efficiency} reports estimated inference FLOPs for LLaMA-3-8B with $T=32$ generated tokens.
Despite requiring two decoding passes, DPS introduces only a modest end-to-end overhead due to shared prompt computation.

Preference Head discovery is performed offline and amortized across users and inference runs, and therefore does not affect deployment-time latency.
Overall, DPS achieves interpretable personalization with inference efficiency comparable to prior contrastive decoding methods.

\subsection{Human Evaluation}
\label{sec:human_eval}

Automatic metrics do not fully capture alignment with user-specific preferences.
We therefore complement them with both human and LLM-based evaluation on LaMP-4, a personalized news headline generation task.

\paragraph{Human Annotation.}
We sample 100 test instances from LaMP-4.
For each instance, annotators are shown the user profile, the input article, and two anonymized candidate headlines produced by CAD and DPS in randomized order.
They are asked to choose the output that better matches the user profile while remaining relevant, fluent, and faithful to the input, and final decisions are obtained by majority vote.

\paragraph{LLM-based Evaluation.}
We additionally use \textbf{GPT-5.2} as an automatic judge.
Using the same blinded pairwise setup, the judge compares two anonymized outputs and assigns a 1--5 score to each output on five dimensions: relevance, fluency, style, preference alignment, and factuality.
We report the average per-method ratings across all evaluated instances.

\paragraph{Results.}
Table~\ref{tab:human_llm_eval} shows that human annotators prefer DPS over CAD in terms of profile alignment on LaMP-4.
The LLM-based evaluation shows a similar trend, with DPS achieving higher scores on style and preference alignment while maintaining strong relevance, fluency, and factuality.

\begin{table}[!t]
\small
\centering
\resizebox{\columnwidth}{!}{
\begin{tabular}{lcccc}
\toprule
\textbf{Evaluator} & \textbf{Metric} & \textbf{CAD} & \textbf{DPS} \\
\midrule
Human & Alignment Win (\%) & 34 & \textbf{40} \\
\midrule
\multirow{5}{*}{GPT-5.2}
& Relevance   & 3.97 & \textbf{4.45} \\
& Fluency     & 4.51 & \textbf{4.83} \\
& Style       & 3.62 & \textbf{3.91} \\
& Alignment   & 3.63 & \textbf{3.93} \\
& Factuality  & 4.08 & \textbf{4.21} \\
\bottomrule
\end{tabular}}
\caption{Human and LLM-based evaluation results on LaMP-4.
Human evaluation reports profile-alignment win rates.
LLM-based evaluation uses GPT-5.2 and reports average per-method ratings on a 1--5 scale.}
\label{tab:human_llm_eval}
\vspace{-0.3cm}
\end{table}

\section{Conclusion}

We present Differential Preference Steering (DPS), a training-free personalization framework grounded in mechanistic interpretability.
By causally identifying Preference Heads and amplifying their contribution at decoding time, DPS enables controllable and interpretable personalization without modifying model parameters.
Experiments show that DPS consistently improves personalization fidelity while preserving content quality and efficiency.
Beyond performance gains, our results provide evidence that user preferences are encoded in sparse, specialized attention heads, offering a mechanistic view of personalization in transformer-based language models.
These findings suggest that personalization can be understood through localized internal components rather than only through black-box adaptation.

\newpage
\section*{Limitations}

While Differential Preference Steering improves personalization fidelity in large language model outputs, it has several limitations.
DPS requires access to internal model components, including attention heads and intermediate activations, which makes it unsuitable for black-box API models.
In addition, DPS introduces additional computation during decoding by requiring a second forward pass with masked Preference Heads.
Although this overhead is modest in practice, it may be undesirable in latency-critical settings.
Future work could explore approximating Preference Head effects in black-box models through lightweight logit biasing or prompt-based steering, as well as further optimizing inference efficiency.
Moreover, because DPS explicitly amplifies user-specific preference signals, it may unintentionally reinforce undesirable biases or overemphasize narrow profile patterns when user profiles are noisy, incomplete, or unrepresentative.

\bibliography{custom}

\appendix

\section{Additional Experimental Details}
\label{sec:appendix_experiments}

\begin{figure*}[!ht]
\centering
\includegraphics[width=\textwidth]{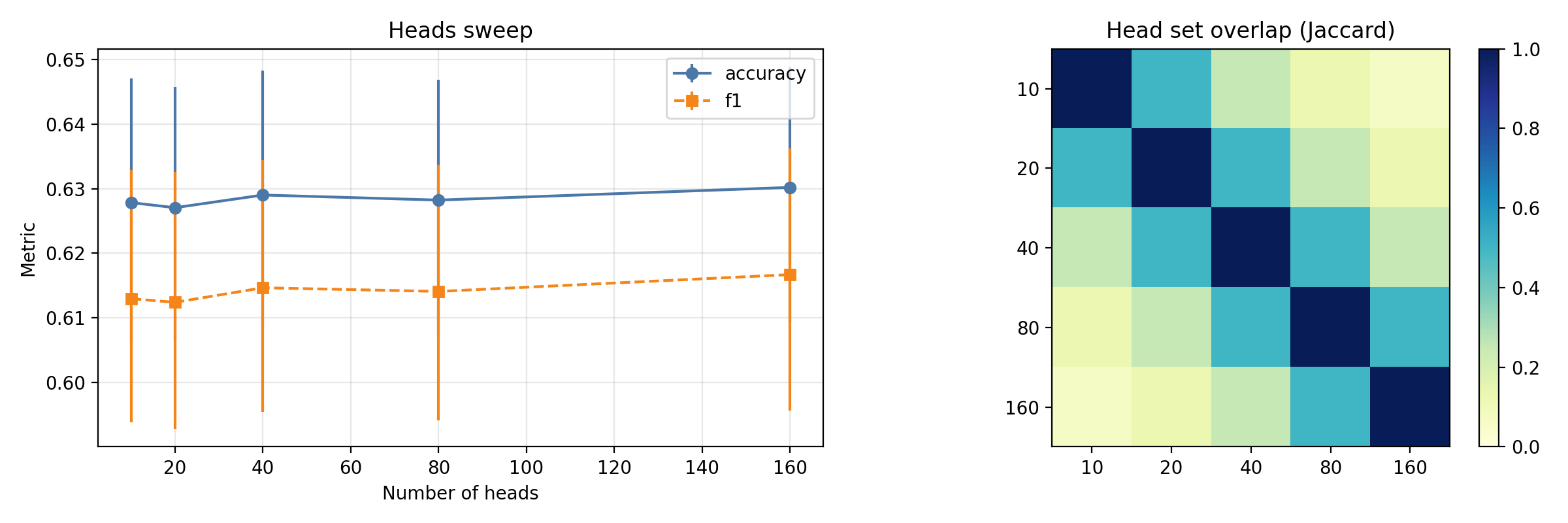}
\vspace{-0.1cm}
\caption{Performance as a function of the number of selected Preference Heads $K$.
Accuracy and F1 scores remain stable across a wide range of $K$, with performance saturating at moderate values, indicating that personalization signals are concentrated in a limited subset of heads.}
\label{fig:heads_sweep}
\vspace{-0.4cm}
\end{figure*}

\begin{figure*}[!ht]
\centering
\includegraphics[width=0.8\textwidth]{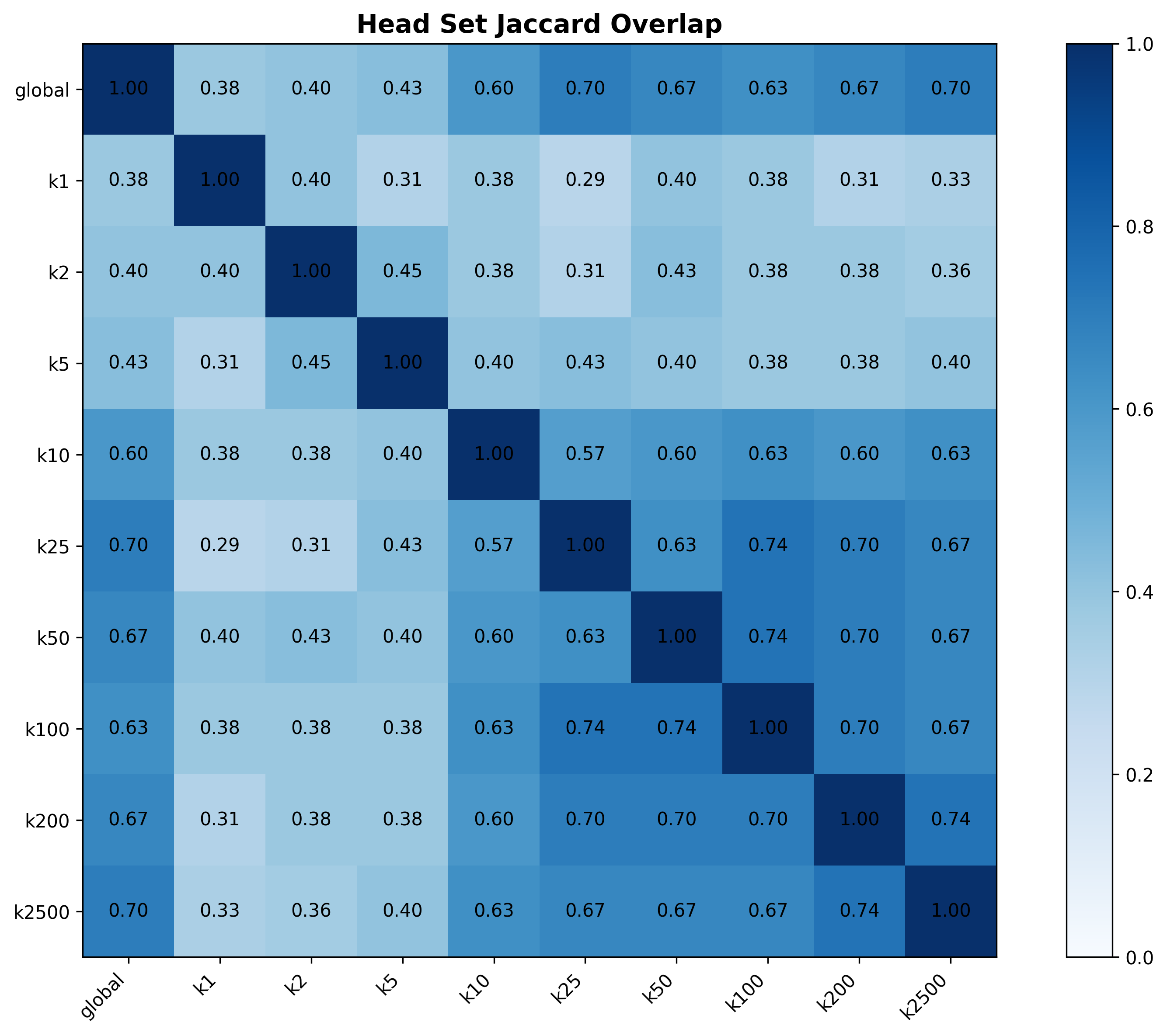}
\vspace{-0.1cm}
\caption{Jaccard overlap between Preference Head sets selected at different values of $K$.
Overlap increases smoothly as $K$ grows, indicating that top-ranked Preference Heads are consistently preserved while additional heads contribute diminishing variation.}
\label{fig:headset_jaccard}
\vspace{-0.4cm}
\end{figure*}

\begin{figure*}[!ht]
\centering
\includegraphics[width=0.9\textwidth]{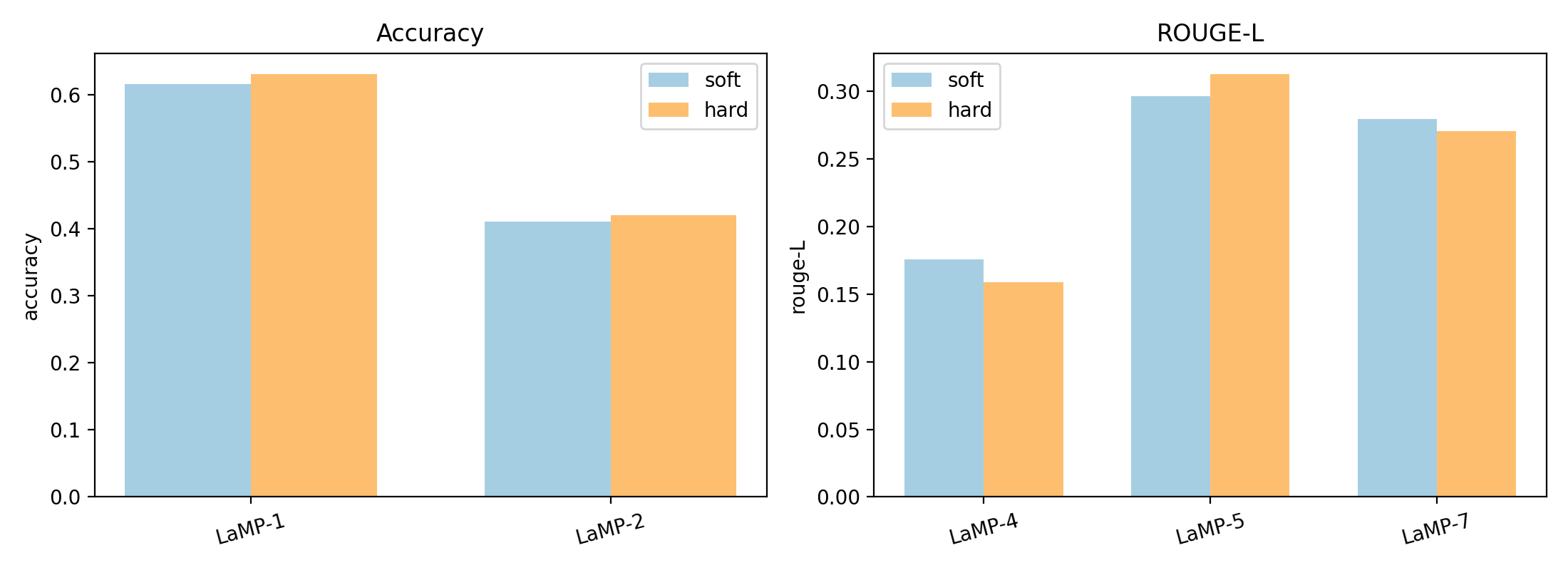}
\vspace{-0.1cm}
\caption{Comparison of hard and soft routing strategies for cluster-aware DPS across LaMP tasks.
Hard routing favors classification tasks, while soft routing yields more consistent improvements on generation tasks.}
\label{fig:routing_ablation}
\vspace{-0.4cm}
\end{figure*}

\subsection{Dataset Details}

We provide additional details of the LaMP benchmark used in our experiments.
LaMP evaluates personalized language modeling by conditioning generation or prediction on user profiles constructed from historical interactions.
Tasks span multiple output formats and supervision types, including free form generation, classification, and regression.

Table~\ref{tab:lamp_stats} summarizes the statistics of the datasets used in this work.

\begin{table}[ht]
\small
\centering
\resizebox{\columnwidth}{!}{
\begin{tabular}{lccc}
\toprule
\textbf{Task} & \textbf{Type} & \textbf{\# Users} & \textbf{\# Instances} \\
\midrule
News Headline Gen. & Gen. & $\sim$1.5K & $\sim$18K \\
Scholarly Title Gen. & Gen. & $\sim$1.2K & $\sim$14K \\
Tweet Paraphrasing & Gen. & $\sim$1.0K & $\sim$12K \\
Citation Identification & Cls. & $\sim$2.0K & $\sim$20K \\
Movie Tagging & Cls. & $\sim$1.8K & $\sim$16K \\
Product Rating & Reg. & $\sim$2.5K & $\sim$25K \\
\bottomrule
\end{tabular}}
\caption{Approximate dataset statistics for LaMP tasks used in our experiments.
Gen. denotes generation, Cls. denotes classification, and Reg. denotes regression.}
\label{tab:lamp_stats}
\vspace{-0.4cm}
\end{table}

\subsection{Baseline Methods}

\paragraph{Context Aware Decoding (CAD)}
CAD improves generation by contrasting logits conditioned on context with a weakened contextual signal.
It has been shown effective for resolving knowledge conflicts and improving contextual faithfulness.

\paragraph{Decoding by Contrasting Layers (DoLa)}
DoLa contrasts logits produced by different transformer layers to amplify factual knowledge encoded in later layers.
It is a representative contrastive decoding method operating over internal model representations.

\paragraph{Decoding by Contrasting Retrieval Heads (DeCoRe)}
DeCoRe identifies retrieval oriented attention heads and applies contrastive decoding to mitigate hallucinations.
Unlike DPS, DeCoRe focuses on factual grounding rather than user specific personalization.

\section{Additional Hyperparameter Analysis}
\label{sec:appendix_hyperparam}

This section reports additional sensitivity analyses for the number of Preference Heads $K$.
We evaluate how varying $K$ affects personalization performance and the stability of the selected head sets across tasks and models.
The results show that DPS is robust over a wide range of settings.
In particular, performance improves rapidly as $K$ increases from small values and saturates at moderate $K$, indicating that personalization is driven by a compact set of heads.
We also observe that the selected head sets remain increasingly stable as $K$ grows, suggesting that top-ranked Preference Heads are consistently preserved while additional heads contribute diminishing returns.

\section{Use of AI Assistants}
\label{sec:appendix_ai_assistants}

We used AI assistants to support limited paraphrasing and language refinement during the writing process.
Specifically, AI tools were used to improve clarity, conciseness, and grammatical correctness of draft text written by the authors.
All technical content, methodological design, experimental setup, analysis, and conclusions were conceived, implemented, and verified by the authors.

\end{document}